\documentclass[runningheads]{llncs}

\usepackage{amsmath,amssymb,amsfonts}
\usepackage{algorithmic}
\usepackage{graphicx}
\usepackage{textcomp}
\usepackage{xcolor}

\begin{document}

\title{cgSpan: Pattern Mining in Conceptual Graphs
}

\author{Adam Faci\inst{1,2} \and
Marie-Jeanne Lesot\inst{1} \and
Claire Laudy\inst{2}}
\authorrunning{A. Faci et al.}
%
\institute{Sorbonne Université, CNRS, LIP6, F-75005 Paris, France \and
LRASC, Thales, 91477 Palaiseau, France}

\maketitle

\begin{abstract}
Conceptual Graphs (CGs) are a graph-based knowledge representation formalism. In this paper we propose cgSpan a CG frequent pattern mining algorithm. It extends the DMGM-GSM algorithm that takes taxonomy-based labeled graphs as input; it includes three more kinds of knowledge of the CG formalism: (a) the fixed arity of relation nodes, handling graphs of neighborhoods centered on relations rather than graphs of nodes, (b) the signatures,  avoiding patterns with concept types more general than the maximal types specified in signatures and (c) the inference rules, applying them during the pattern mining process. The experimental study highlights that cgSpan is a functional CG Frequent Pattern Mining algorithm and that including CGs specificities results in a faster algorithm with more expressive results and less redundancy with vocabulary.

\keywords{Conceptual Graphs \and Frequent pattern mining.}
\end{abstract} 

\section{Introduction}
Conceptual Graphs (CGs) \cite{chein_conceptual_2008} represent knowledge as graphs containing concept nodes and relation nodes which refer to ontological knowledge in vocabulary. In their simplest form, they are similar to taxonomy-based labeled graphs (TLGs) i.e. labeled graphs with a partial order defined on the set of labels corresponding to an \textit{is-a} hierarchy. While there are Frequent Pattern Mining (FPM) algorithms considering TLGs, an FPM algorithm taking CGs as input has not yet been proposed to the best of our knowledge.
Yet mining patterns from sets of structures is a prevalent research subject
\cite{yan_gspan_2002,han_frequent_2007,iyer2018asap,elseidy2014grami}. Taking TLGs as input has been the work of a few propositions \cite{inokuchi_mining_2004,cakmak_taxonomy-superimposed_2008,petermann_mining_2017} and we consider them as a basis to design a CGs pattern mining algorithm.

We propose cgSpan based on these algorithms. We consider CGs as TLGs with more layers of information in order to reuse an existing algorithm of the state of the art. We propose to exploit three differences with the TLGs model; the relations fixed arity, the signatures and the inference rules. The first difference corresponds to the biparticity of CGs, the second one to the constraints on concept nodes labels connected to relation nodes and the third one to deduction mechanisms.

This paper has two goals. First it aims at defining a functional frequent pattern mining algorithm running on a CG database, taking the specificities and the additional knowledge in CGs into account, as compared to TLGs. Second it aims at showing that using such specificities results in a more efficient algorithm in memory space, speed and quality of output as compared to existing algorithms on TLGs.

Section~\ref{sec:edla} presents conceptual graphs as well as the current state of the art on taxonomy-based labeled graph pattern mining algorithms. Section~\ref{sec:proposition} describes the proposed algorithm named  cgSpan. Section~\ref{sec:expe} describes the experimental study and the considered quality criteria, as well as the obtained results. Section~\ref{sec:conclusion} concludes on this work and discusses some directions for future works

\begin{figure}[t]
\centerline{\includegraphics[width=0.5\linewidth]{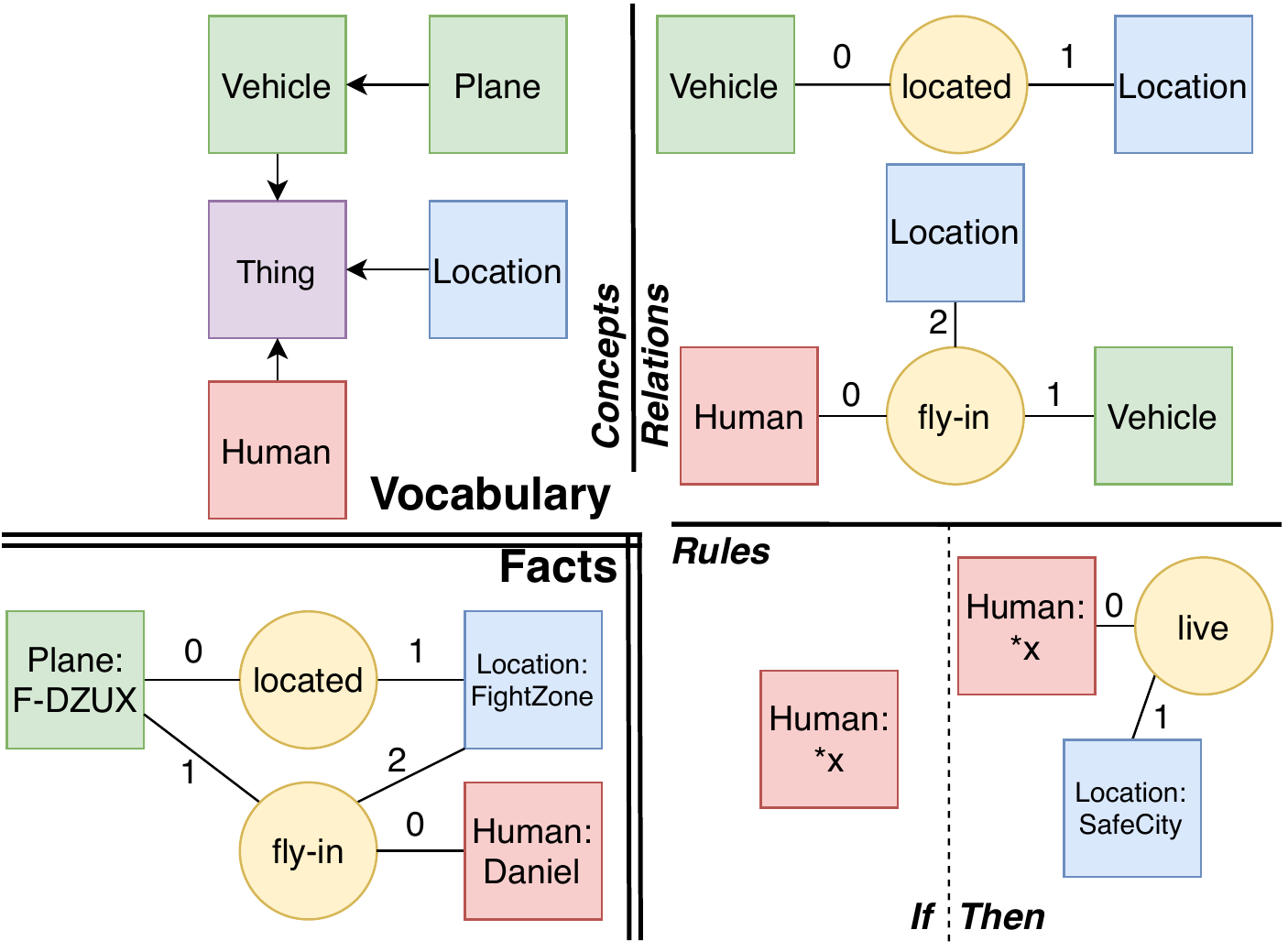}}
\caption{Example of a CG with its vocabulary: CG in the bottom left part, a hierarchy on concepts in top left part, two signatures in the top right part and an inference rule in the bottom right part.}
\label{fig:example}
\end{figure}

\section{State of the Art}\label{sec:edla}
{This section briefly recalls the definitions of conceptual graphs and summarizes some graph pattern mining algorithms.}

\subsection{Conceptual Graphs} \label{sub:CG}
{Conceptual graphs \cite{chein_conceptual_2008} are a family of formalisms of knowledge representation, made of ontological and factual knowledge. They define sets of bipartite graphs, the CGs, representing the facts, all referring to ontological knowledge stored in the so-called vocabulary; an example is illustrated on Fig.~\ref{fig:example} and commented below. Several levels of expressiveness have been defined, see \cite{chein_conceptual_2008} for an exhaustive review, the formal definitions of the notions used in the paper are recalled below. 
}

The ontological part of a CG is a \textit{vocabulary}, defined as a {5-tuple\textit{ V = $(T_C,T_R,\sigma,I,\tau)$}}. $T_C$ and $T_R$ that respectively correspond to concept and relation types are two partially ordered disjoint finite sets, where ordering corresponds to generalisation. $T_C$ is illustrated in the top left part of Fig.~\ref{fig:example}. It contains a greatest element $\top$, represented as "Thing" in Fig.~\ref{fig:example}. $T_R$ is partitioned into subsets $T^1_R\ldots T^k_R$,  $1\ldots k$ ($k \geq 1$) respectively, meaning that each relation type has an associated fixed arity. $\sigma$ is a mapping associating a signature, illustrated on top right part of Fig.~\ref{fig:example}, to each relation. $I$ is a set of individual markers. $\tau$ is a mapping from $I$ to $T_C$.

The factual part are the CGs themselves. A CG is a 4-tuple \textit{G = (C,R,E,label)}. $G$ is a bipartite labeled multigraph as illustrated on the bottom left part of Fig.~\ref{fig:example}. $C$ and $R$ correspond to concept and relation nodes, where  elements of $C$ are pairs from $T_C\times~I$ and elements of $R$ are elements of $T_C$. $E$ contains all the edges connecting elements of $C$ and $R$ and $label$ is a labelling function.

An extension to the previous basic CG setting consists in defining so-called inference rules, that define deduction principles to complete CG, either including additional nodes or updating their labels. As 
illustrated in bottom right part of Fig.~\ref{fig:example}, such a $\lambda$-rule is an ordered pair of $\lambda$-CG made of a hypothesis and a conclusion, where a $\lambda$-CG is a CG with defined connection nodes, i.e. nodes with generic marker with an associated variable used to match nodes from hypothesis to nodes from conclusion. Fig.~\ref{fig:example} illustrates the two connection nodes matching with the associated variable "$*x$". The application of a rule can be the extension of a CG, the conclusion then including the hypothesis, or the specialization of a pattern, the conclusion then being a copy of the hypothesis with more specific labels, but it can be both. Fig.~\ref{fig:example} illustrates the case of an extension rule. Formal definitions can be looked up to in \cite{chein_conceptual_2008}.

We base our proposition on algorithms taking taxonomy-based labeled graphs (TLGs) \cite{petermann_mining_2017} as input. A taxonomy is a set of labels with a partial order relation. For instance, the hierarchy of concepts in Fig.~\ref{fig:example} is a taxonomy. The partial order relation is the generalisation, meaning that if $A$ generalises $B$, if any instance of $B$ is an instance of $A$. A TLG is a labeled graph whose labels are part of a taxonomy, less expressive than a CG. Fig.~\ref{fig:example} would represent a TLG if there were no distinction between concept and relation nodes, no relation fixed arity, no signatures, no inference rules and no individual markers.

\subsection{Subgraph Mining}\label{sub:edlamining}

{Frequent pattern mining (FPM) algorithms, e.g. summarized in~\cite{han_frequent_2007}, are central in the data mining community.} They take a database as input and return a set of frequent patterns, by counting support, i.e. the number of occurrences, of each candidate, i.e. a potential frequent pattern, in the considered data. 
Patterns can be extracted from  sets, also called transactions, sequences or  structures for example. 
In the case of complex instances such as sequences or graphs, the support can be defined as the number of instances containing the candidate, not considering multiple occurrences within an instance as a weight. In this paper, we consider the classic choice where a graph supports a pattern if and only if there exists a homomorphism from the pattern to a subgraph of the considered graph.

One of the most used graph pattern mining algorithms is gSpan   \cite{yan_gspan_2002}, that applies to labeled graphs and relies on encoding graphs as  sequences, where a unique sequence corresponds to a unique graph, switching the context as well from subgraph mining to subsequence mining. It has been extended to the case of taxonomy-based labeled graphs \cite{inokuchi_mining_2004,cakmak_taxonomy-superimposed_2008} and then further enriched to add the possibility of mining directed TLGs with many taxonomies \cite{petermann_mining_2017}. 

These TLGs pattern mining algorithms add the inclusion of hierarchies to find more patterns and gain in efficiency. Indeed, label generalization leads to a more lenient label comparison, as two different labels can be generalized to a similar one, increasing the number of returned patterns. Also, they first mine for structural patterns before considering labels by generalizing to the most general type, resulting in more patterns explored at once. They also tackle the problem of the massive amount of patterns returned in classic pattern mining approaches by pruning or not exploring irrelevant patterns. The pattern relevancy is determined by several measures such as  over-generalization and statistical significance~\cite{petermann_mining_2017}.

Our proposition is based on the DMGM-GSM algorithm  (Directed MultiGraph Miner - Generalized Subgraph Mining) \cite{petermann_mining_2017}. The latter takes into account the taxonomy on labels  in 3 steps: first for each node in a TLG, the label is replaced by the path to its upmost type. Then patterns are mined by considering only the top type in the newly encoded taxonomy paths labels. Finally, retrieved patterns are successively specialized until they are no more frequent, i.e. with a support count lower than the defined support threshold.

\section{{cgSpan: a CG Frequent Pattern Mining Algorithm}}\label{sec:proposition}
cgSpan is the first frequent pattern mining method running on Conceptual Graphs, to the best of our knowledge. Based on DMGM-GSM \cite{petermann_mining_2017}, it is conceived as taking consecutive enrichment steps from TLGs to CGs into account. We use the additional information in CGs to restrict the output and ensure both conformity of result and speed. 

\subsection{Overview}

cgSpan is made of three modules, each one being dedicated to one specificity of CGs as opposed to TLGs: the rules fixed arity, the signatures and the inference rules. They can be combined to exploit simultaneously all characteristics of CGs, as illustrated in Section~\ref{sec:expe}. They are described in turn in Sections~\ref{sub:neighborhood} to~\ref{sub:rules}, this section describes their common points. 

The essential part of cgSpan is a pre-processing step which translates each CG of the input database to TLG. The basic principle is similar to that of DMGM-GSM: labels are built by concatenating all concept types, from the top type to the node type, separating them with a '\_'. For instance "Plane:F-DZUX" from Fig.~\ref{fig:example} is replaced by "Thing\_Vehicle\_Plane\_F-DZUX". Then this step is modified accordingly by each of the three modules to take into account CG specificities, especially by the neighborhood module (see Section \ref{sub:neighborhood}).

Then the pattern mining is run with DMGM-GSM, only modified by the inference rule module described in Section~\ref{sub:rules} to increase efficiency.

Finally the post-processing translates the obtained frequent patterns back to the CG formalism. As described in Section~\ref{sub:sign}, the signature module includes an additional pruning step.

\subsection{Exploiting Relation Arity and Neighborhood Nodes}\label{neigh}
\label{sub:neighborhood}

The first specificity of CGs we propose to exploit is the arity of the relation nodes: any relation node is necessarily connected to a known number of concept nodes. In this regard, when operating on candidates, any relation node addition should immediately result in the addition of its connected nodes.

\begin{figure}[t]
\centerline{\includegraphics[width=0.8\linewidth]{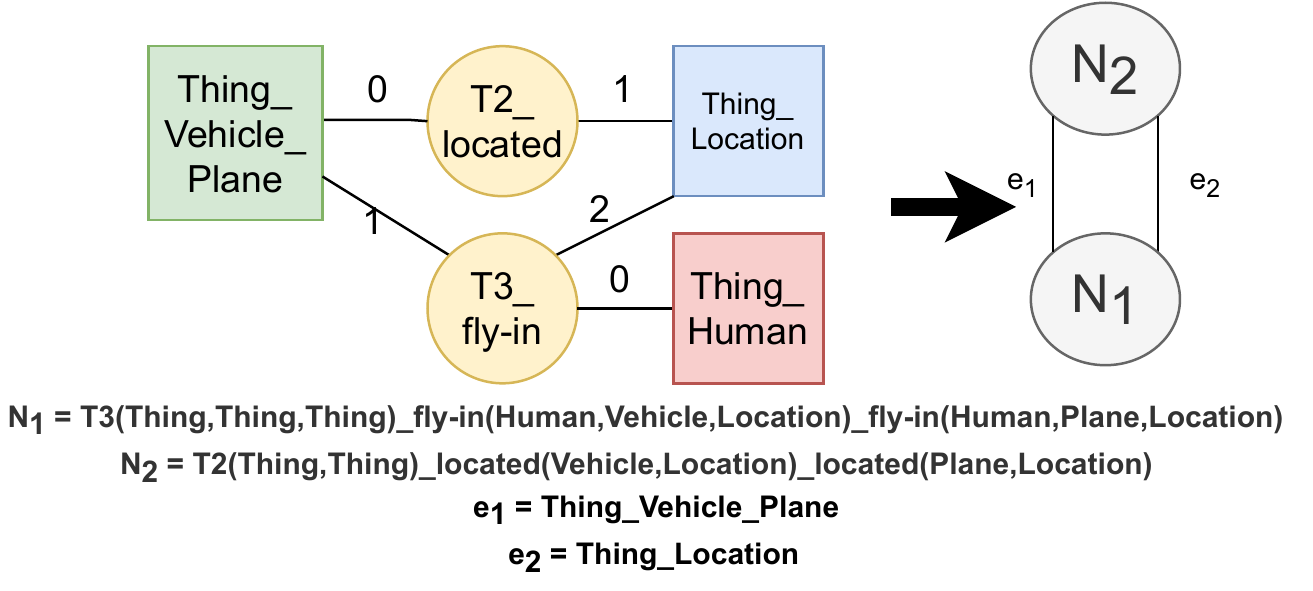}}
\caption{Translation from CG to TLG as sets of bricks. The taxonomy-path labels detailed below start from top types and specialize all node labels at the same time.}
\label{fig1}
\end{figure}

We therefore define an \emph{elementary brick} as a node and its connected nodes of the input CGs, so as to avoid partial neighborhoods in returned patterns and to increase efficiency. Fig.~\ref{fig1} illustrates how a CG can be encapsulated as a TLG of elementary bricks where "T3" is a relation type more general than "fly-in".

A brick is associated to a single TLG node. Its label is defined as follows: it is the concatenation of the taxonomy-enriched labels of the relation node and each of its associated concept nodes, in the order specified in CGs by the edge labels. Finally, edges are built between bricks that share a common concept node, as illustrated in the right part of Fig.~\ref{fig1}. The edge labels are the taxonomy paths of the bricks common concept nodes. Fig.~\ref{fig1} shows for instance the transformation from a brick in a CG with "A plane is flown by a human in a location" to $N_1$.

\subsection{Exploiting Signatures}\label{sub:sign}

The second specificity of CGs considered in cgSpan is the signatures. They define for each relation type a restriction on concept types: for each relation node, they specify a maximal generic type for each connected concept node. We propose to exploit this information in the pre-processing and post-processing steps, respectively to restrict the label generalization of concept nodes and to prune patterns to avoid redundancy with signatures.

\begin{figure}[t]
\centerline{\includegraphics[width=0.8\linewidth]{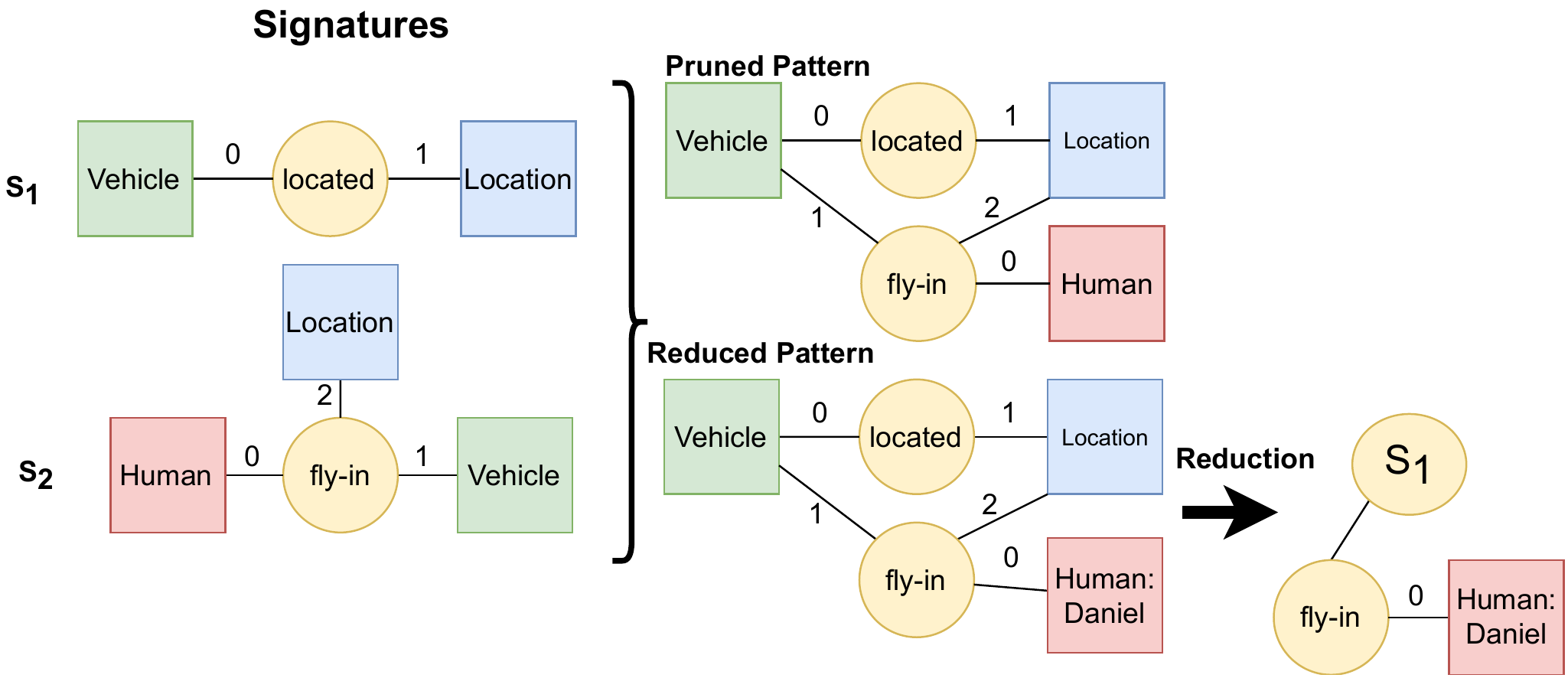}}
\caption{Pattern pruning in post-processing step when considering signatures. The top right pattern is pruned as it is a mere aggregation of signatures. The bottom right pattern is reduced by replacing the redundant part to $S_1$ with a reference.}
\label{fig2}
\end{figure}

As an instance we consider in this section the relation "fly-in" with  signature $(Human, Vehicle, Location)$ as illustrated in Fig.~\ref{fig2}.

In the pre-processing step we propose to prune the taxonomy path, not building it up to the most general type, but up to the level indicated in the signature. For each relation node in a CG, concept nodes connected to this relation see their labels compared to the corresponding signature. Each comparison is then followed by a cut of the taxonomy path down to the matching type in signature. 
For instance, "fly-in(Thing\_Human,Thing\_Vehicle\_Plane,Thing\_Location)" is replaced by "fly-in(Human,Vehicle\_Plane, Location)" in Fig.~\ref{fig2}.

In the post-processing step, two kinds of patterns are affected. 
First, we propose to remove patterns that can be reduced to a set of connected signatures.
Second, we propose to modify patterns including parts that can be reduced to a set of connected signatures by replacing the latter by references to signatures.

For instance, "fly-in(Human,Vehicle,Location)" is ignored while "fly-in(Human, Vehicle\_Plane, Location)" is not, since "Plane" is a specialization of "Vehicle" type specified in signature as illustrated in Fig.~\ref{fig2}.

\subsection{Exploiting Rules}\label{rules}
\label{sub:rules}

{The third specificity taken into account by cgSpan is positive rules, considering the ones including extension and the ones including specialization.}

{A positive rule including specialization specifies that when a CG includes the rule hypothesis, some nodes specified in rule conclusion can be specialized.}
{During pre-processing, after the replacement by taxonomy paths in CGs, we propose to extend this path for each node matching such a rule to the type in conclusion.}

{On the contrary, a positive rule including extension enables the extension of CGs in database.}
{During the pattern mining step, we propose to extend all patterns matching the rule hypothesis to the rule conclusion. In the process, the rule hypothesis and intermediate patterns are ignored.
It can be applied more than once, thus extending a matching CG at each application, but we limit their use to one.}

\begin{figure}[t]
\centerline{\includegraphics[width=0.8\linewidth]{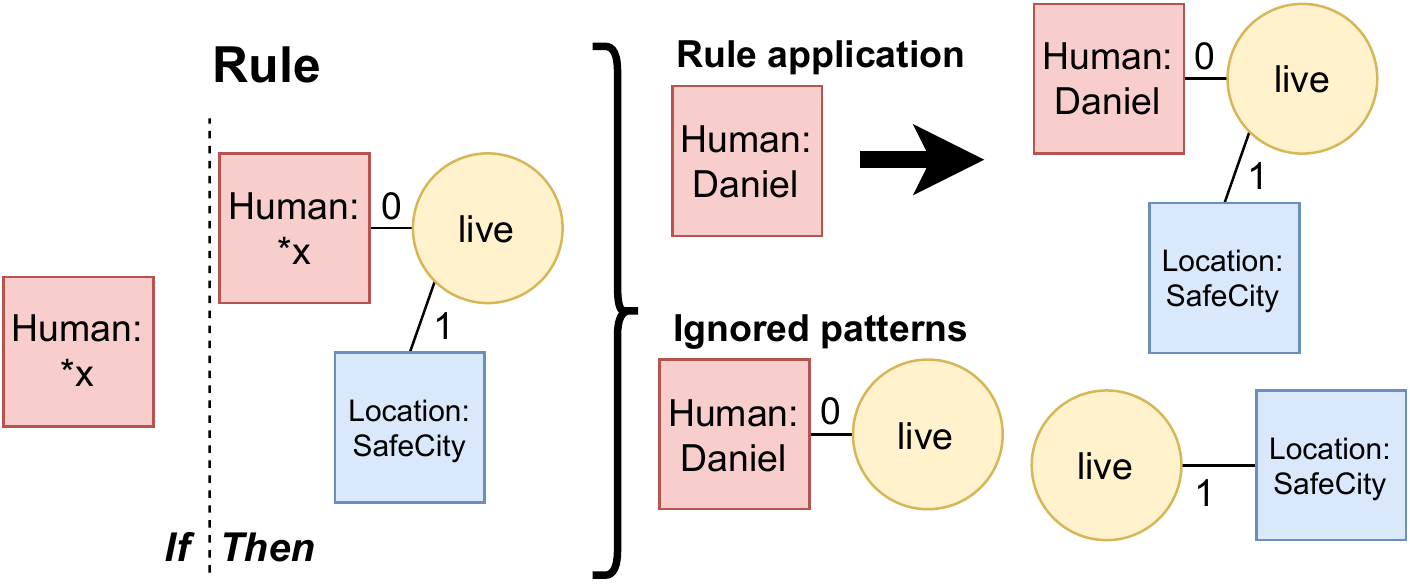}}
\caption{Rule application example and examples of patterns not explored using this technique. The rule is only applied once on the top right part. Ignored patterns are potentially not explored if candidates start with a single concept node with type "Human".}
\label{fig3}
\end{figure}

Fig.~\ref{fig3} illustrates the application of the rule from Fig.~\ref{fig:example}. All intermediate patterns, from hypothesis to conclusion, are not explored when using such rules since the extension is in one step.

\section{Experimental Study}\label{sec:expe}
{This section describes the experiments we conducted, presenting the considered synthetic data and quality criteria and discussing the obtained results.}

\subsection{Data Generation}
{To the best of our knowledge, no CG dataset of quality is available. Existing ones are either extremely simple or do not fully respect the CGs formal restrictions. One possibility would be to use a translation method such as $T_{nat}$ \cite{baget_translation_2010} that makes it possible to obtain CG datasets from real datasets. However it usually results in datasets with mainly factual or mainly ontological knowledge. Moreover it does not allow to define the expected results in terms of frequent patterns to identify in the data, making it difficult to validate the proposed algorithm. Consequently we propose CGGen, a new algorithm to generate CG datasets from a set of constraints, either from a set of defined frequent patterns, used as components of generated CGs, or from frequency distributions that the generated CGs need to respect.} CGGen thus enables the definition of expected results, as expected frequent patterns on one hand and frequency distribution on the other hand, that can be used to validate our algorithm.

{We use two datasets in the experimental results, denoted $D_1$ nad $D_2$, consisting in a few hundreds of nodes.} They consist of 1000 graphs of around 30 nodes each. The hierarchies contain 50 concept types and 20 relation types.  $D_2$ CGs follow two distributions used for generation: one over their size and one over their labels.

{$D_1$ enables the definition of expected results. First we expect the set of predefined CGs used as seed to generate these datasets to be present in the frequent patterns. Then we expect that some simple constraints verified by~$D_1$ are also verified by the set of frequent patterns. Also we expect that the distribution on labels in $D_2$ is followed by the returned patterns.}

\subsection{Criteria}\label{sub:criteria}

The comparison of the cgSpan returned patterns with the expected ones allows to make use of the classic precision and recall criteria, respectively defined as the proportion of expected patterns present in the returned ones, and reciprocally. 

In addition, to assess the computational efficiency of the proposed cgSpan algorithm, the redundancy of patterns is assessed, defined as the proportion of pruned patterns w.r.t. all patterns, returned or pruned. The greater it is, the more redundancy has been removed, the better it is.

Finally, to assess efficiency, the run time of cgSpan is compared to that of DMGM-GSM.

\subsection{Experimental Results}

We process $D_1$ and $D_2$ translated to TLG with DMGM-GSM and process them with four variants of cgSpan, considering each module, as described in Sections~\ref{sub:neighborhood}, \ref{sub:sign} and \ref{sub:rules}, individually and full cgSpan that includes all three modules.

Tab.~\ref{tab1} gives the results obtained for $D_1$. All algorithms retrieve all expected patterns or combination of expected patterns {so all versions attain the functional goal} and obtain a recall equal to 100\%. Regarding precision, all variants not taking elementary bricks into account return partial neighborhoods, reducing their performance for this criterion. The use of signatures prunes some of these incomplete patterns since most of them contain top concept nodes. We can spot that an increase in redundancy is correlated with an increase in precision. It can be interpreted as "Patterns that are pruned are mostly not expected patterns". Finally, the use of elementary bricks seems to increase the time efficiency significantly. Indeed, there is a gain of almost 20\% in time while the gain with the use of rules is less than 10\%. These observations have been confirmed by experiments with other generated databases with varying parameters. Tests on a huge database with millions of nodes are included in future works to observe how well this improvement keeps up with the increase in size.

\begin{figure}[t]
\centerline{\includegraphics[width=0.8\linewidth]{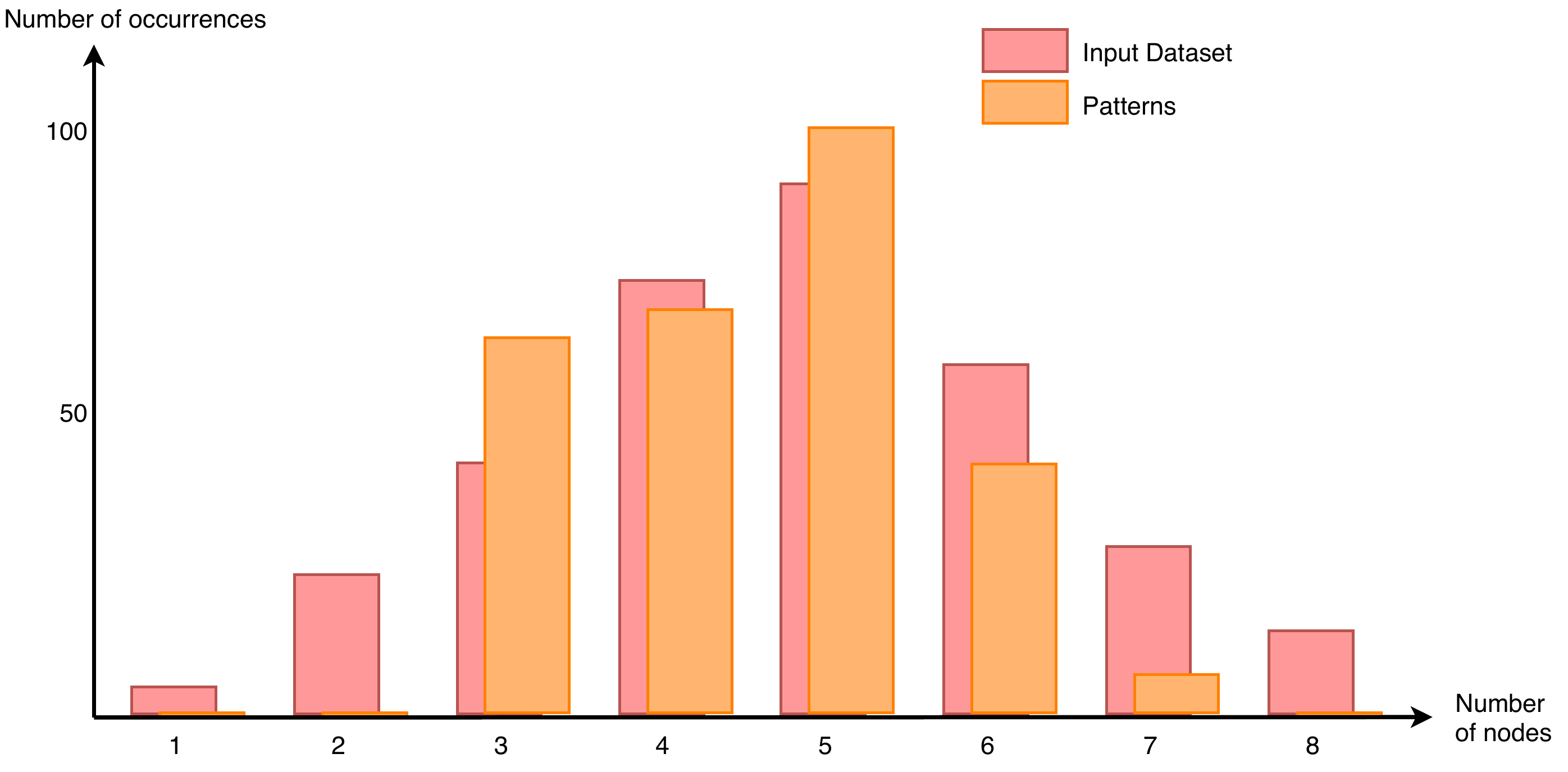}}
\caption{Frequency distributions over $D_2$ and over the set of returned frequent patterns with cgSpan.}
\label{fig4}
\end{figure}

Fig.~\ref{fig4} presents the results with cgSpan on $D_2$, showing the number of occurrences of the expected and output patterns depending on their size. It shows the frequency distribution on the number of occurrences of patterns returned by cgSpan according to their size. The frequency distribution over the input dataset has been used to generate $D_2$, so we use both distributions to analyze results.
Note that when counting patterns, only the maximal patterns, i.e. patterns not included in another pattern, contribute to this illustration: all counted patterns of size 3 are not present in the counted patterns of greater size.
We can observe a global correlation between the input dataset and the frequent patterns. The fact that there is no frequent pattern of size 1 or 2 is explained with the elementary brick restriction that does not allow partial neighborhood while in $D_2$ these CGs correspond to isolated concept nodes.

\begin{table}[t]
\caption{Comparison of cgSpan and its variants with $D_1$}
\begin{center}
    \begin{tabular}{|c|c|c|c|c|c|}
        \hline
        \textbf{Test}  & \textbf{Rec. (\%)} & \textbf{Prec. (\%)} & \textbf{Red. (\%)}& \textbf{T-Eff. (\%)}\\
        \hline
        {DMGM-GSM}  & 100 & 57 & 0 & 100\\
        \hline
        cgSpan with bricks & 100 & 83 & 0 & 83\\
        \hline
        cgSpan with signatures & 100 & 75 & 38 & 97\\
        \hline
        cgSpan with rules & 100 & 63 & 20 & 93\\
        \hline
        Full cgSpan & 100 & 85 & 46 & 79\\
        \hline
    \end{tabular}
    \label{tab1}
\end{center}
\end{table}

\section{Conclusion and Future Works}\label{sec:conclusion}

This paper presents cgSpan, an algorithm to mine frequent patterns from a CG database. The particularities of CGs are exploited to produce a functional algorithm, and quality of result is increased w.r.t. defined criteria. We use the concept and relation difference and inference rules to speed up the process and we use signatures and other elements to prune less relevant patterns.

Future works will aim at extending the proposed cgSpan algorithm, considering even richer variants of CGs, in particular nested CGs \cite{chein_conceptual_2008}. Another direction will focus on the definition of quality criteria: it may be interesting to exploit statistical significance such as on based on Chung–Lu model \cite{chung2002average} adapted to CGs to propose a functional statistical significance criterion for CGs mining algorithms such as cgSpan, used both to improve results and validation.

\bibliographystyle{splncs04}
\bibliography{bib}

\end{document}